\documentclass{article}
\usepackage{arxiv}
\usepackage[utf8]{inputenc} 
\usepackage[T1]{fontenc}    
\usepackage{hyperref}       
\usepackage{url}            
\usepackage{booktabs}       
\usepackage{amsfonts}       
\usepackage{nicefrac}       
\usepackage{microtype}      
\usepackage{graphicx}

\title{HARK Side of Deep Learning - From Grad Student Descent to Automated Machine Learning}

\author{
  Oguzhan~Gencoglu\\
  Top Data Science Ltd.\\
  Helsinki, Finland \\
  \texttt{oguzhan.gencoglu@topdatascience.com}
  \And 
  Mark~van~Gils \\
  VTT Technical Research Centre of Finland Ltd.\\
  Tampere, Finland \\
  \texttt{mark.vangils@vtt.fi}
  \And 
  Esin~Guldogan \\
  Huawei Technologies\\
  Tampere, Finland \\
  \texttt{esin.guldogan@huawei.com}
  \And
  Chamin~Morikawa \\
  Morpho Inc. \\
  Tokyo, Japan \\
  \texttt{c-morikawa@morphoinc.com}
  \And 
  \vspace{3.9 mm}
  Mehmet~S\"uzen \\ 
  J\"ulich, Germany \\
  \texttt{suzen@acm.org}
  \And 
  Mathias~Gruber \\
  Novozymes \\
  Copenhagen, Denmark \\
  \texttt{mafg@novozymes.com}
  \And 
  Jussi~Leinonen \\
  Bayer\\
  Espoo, Finland \\
  \texttt{jussi.leinonen@bayer.com}
  \And
  Heikki~Huttunen \\
  Tampere University \\
  Tampere, Finland \\
  \texttt{heikki.huttunen@tuni.fi}
} 

\begin{document}
\maketitle

\begin{abstract}
Recent advancements in machine learning research, i.e., deep learning, introduced methods that excel conventional algorithms as well as humans in several complex tasks, ranging from detection of objects in images and speech recognition to playing difficult strategic games. However, the current methodology of machine learning research and consequently, implementations of the real-world applications of such algorithms, seems to have a recurring \textit{HARKing} (Hypothesizing After the Results are Known) issue. In this work, we elaborate on the algorithmic, economic and social reasons and consequences of this phenomenon. We present examples from current common practices of conducting machine learning research (e.g., avoidance of reporting negative results) and failure of generalization ability of the proposed algorithms and datasets in actual real-life usage. Furthermore, a potential future trajectory of machine learning research and development from the perspective of accountable, unbiased, ethical and privacy-aware algorithmic decision making is discussed. We would like to emphasize that with this discussion we neither claim to provide an exhaustive argumentation nor blame any specific institution or individual on the raised issues. This is simply a discussion put forth by \textit{us}, insiders of the machine learning field, reflecting on \textit{us}.  
\end{abstract}


\section{Introduction}
\textit{Hypothesizing after the results are known} (HARKing)~\cite{kerr1998harking} occurs when researchers masquerade one or more post hoc hypotheses as \textit{a priori} hypotheses. This means that instead of following a traditional hypothetico-deductive model~\cite{hempel1966}, in which previous knowledge or conjecture is used to formulate hypotheses that are then tested, the researcher instead looks at the results first and then forms a \textit{post hoc} hypothesis. HARKing can occur in different forms, such as \textit{constructing}, \textit{retrieving} or \textit{suppressing} hypotheses after the results are known~\cite{kerr1998harking}. A number of studies in recent years have examined and discussed the incidences, causes and implications of such practices within various fields such as management, psychology as well as natural sciences~\cite{murphy2017harking,hitchcock2004prediction,rubin2017does}. 

In recent years, deep learning (DL) methods have dramatically improved the state-of-the-art (SotA) within the fields of speech recognition, visual object recognition, machine translation and several other domains such as drug discovery and genomics~\cite{lecun2015deep}. However, there are certain troubling trends in the current machine learning (ML) research, outlined in~\cite{lipton2018troubling} as failure to distinguish between explanation and speculation, use of mathematics that obfuscates rather than clarifies, and misuse of language. Unfortunately, HARKing has also been one of those recurrent trends in machine learning and especially in deep learning research. Since much of such research is being eagerly applied to real-world applications in both industry and society, such issues are of utmost importance due to the wide impact of machine learning products and services across all walks of life. Transparent and reliable practices are critical when trying to combat suspicions towards new technologies, and the trust needs to be built over long period of time; as acknowledged recently even on the European Commission level~\cite{EC_Communication_2019}.

Our hypothesis is that the recent explosion of advances within the fields of machine learning and in particular deep learning, as well as the hyper-competitive nature of these fields, may potentially be a dangerous breeding ground for various HARKing behaviors, the implications of which are not yet fully explored. At the very least, concerns regarding such behaviours deserve to be critically discussed from different angles so as to encourage best practices when building ML systems and algorithms. It is noted that these issues are not new by themselves. In fact, since as long as data-driven approaches and learning systems have been around, it has been critical, and sometimes difficult, to remain fully objective in analyzing results. Issues have been reported earlier for example, such as self-deception practiced by scientists; finding patterns that are not there~\cite{nuzzo2015fooling,bailey2014a}. 

In this paper we discuss HARKing behavior from different angles: 
\begin{itemize}
  \item Section \ref{sec2} - Competitiveness in DL research leading to questionable improvements of state-of-the-art and claims of novelty.
  \item Section \ref{sec3} - Pressure to create reports that are favorable for publication and aversion towards negative results.
  \item Section \ref{sec4} - The belief that current training datasets are representative of real-world samples.
  \item Section \ref{sec5} - Automated machine learning
  \item Section \ref{sec6} - Explainability, ethics, reproducibility and more for AI systems.
\end{itemize}



\section{Grad Student Descent and SotA-hacking} \label{sec2}
In a typical deep neural network model there are numerous design choices, i.e., tunable parts such as model architecture and hyper-parameters, that affect the predictive performance. Proposing a decent set of these design choices that will result in high generalization ability (relative to the other sets of choices) is difficult mainly due to two reasons. Firstly, due to the inherent non-deterministic and highly non-linear nature of neural networks, it is not trivial to deduce explicit relationships neither between the hyper-parameters and the model performance, nor between the interactions of hyper-parameters themselves. For instance, a large \textit{batch size} is key to speed up neural network training in large distributed computation infrastructures, however, significant degradation in model performance has been observed in practice when large batch sizes are employed~\cite{keskar2016large}. To overcome this issue, typically, hyper-parameters belonging to the \textit{optimizer} need to be tuned. Secondly, as the parameter search space increases exponentially, it is not feasible to apply exhaustive or brute-force search methods. Therefore, a significant portion of deep learning research has been focusing on engineering efficient model architectures and hyper-parameters for specific tasks. 

Even though this manual discovery process has been successful for several applications (often empirically), there has been significant divergence from the traditional hypothesis-driven scientific approach in the methodology of such studies. Instead of hypothesis-forming based on theory, extensive research on previous studies and/or reflection against the existing domain knowledge, \textit{grad student descent} (a cheesy pun referring to the well-known \textit{gradient descent} algorithm) is applied. 

Grad student descent is a type of optimization scheme in which the task of model architecture or hyper-parameter search is assigned to several graduate students, usually to be performed by trying what works and what does not. This is an iterative approach, where one starts with a baseline architecture or possibly with an earlier SotA, measures its performance and applies various modifications by trial-and-error, without a sound hypothesis. Once marginal improvements are observed, iterations of modifications continue further in that direction until a local optimum (often a publishable result) is reached and an explanation is forged. In essence, this whole process is driven by HARKing. Furthermore, this process is performed with a limited set of data, that is used and re-used again and again to find the "optimal" solution (further discussed in Section \ref{sec4}). Oftentimes, final testing on a completely independent test set that has not been touched or observed at all at any moment is not performed and cross-validation is either not used or used under problematic assumptions and/or executions such as performing model tuning and estimation of model error at the same time~\cite{hastie2009elements,cawley2010over}.

The abovementioned HARKing pattern, consequently, results in increased difficulty in distinguishing and identifying why a proposed method works or not. Lack of thorough hypothesis forming prior to experimentation often leads to negligence of comprehensive discussions on the results as well, especially when accompanied with comparison of a single score or metric. For instance, a recent work by Reimers and Gurevych shows that reporting a single performance score is insufficient to compare non-deterministic approaches such as neural networks~\cite{reimers2017reporting}. Their study demonstrates that the seed value for the random number generator can result in statistically significant differences in performances of state-of-the-art methods~\cite{reimers2017reporting}. 

The negative effects of HARKing are not specific to deep learning research alone, and they can be observed in research dealing with traditional machine learning methods as well. However, as the concept of \textit{state-of-the-art} (a method or a set of methods that outperforms all the previously proposed methods for a given machine learning task in a certain metric such as test accuracy, inference speed, training speed etc.) has been disproportionately promoted in DL, both in academy and industry, presence of HARKing is becoming more likely to be overlooked especially if there are claims of advancing the SotA. This phenomenon has been promoting the concept of \textit{SotA-hacking} and publishing of \textit{marginally SotA} results without in-depth analysis or discussion, similar to \textit{p-hacking}, \textit{data dredging} and prevalence of \textit{marginally significant} results in several other fields~\cite{selvin1966data,head2015extent,olsson2019prevalence}. Typical examples of misleading comparisons leading to unfair or inadequate SotA claims include usage of additional training data (the common concept of \textit{transfer learning} in DL), usage of data augmentation, comparison to poorly implemented baselines or ensembling of several models. Similar unjustified claims can be observed in "novelty" of proposed methods as well.

\section{Chronic Allergy to Negative Results} \label{sec3}
\textit{Publication bias}, the phenomenon occurring when the probability of a scientific study being published is not independent of its results~\cite{sterling1959publication}, leads to systematic difference in the findings of published tests of a claim from the findings of all tests of the same claim~\cite{song2000publication}. Often recurring as a \textit{positive outcome bias}, this phenomenon has been observed in several research fields for a long time~\cite{easterbrook1991publication,moscati1994positive,sterling1995publication}. For example in clinical research, studies finding no difference between the study groups were less likely to be published than those with statistically significant results~\cite{easterbrook1991publication}. In fact, there is evidence of negative results being less likely to be published even if they provide corrections of errors in previous studies~\cite{matosin2014negativity}. A similar troubling trend has been prevalent in ML/DL research and arguably HARKing exacerbates this further.

Publishing a null or negative result in the current ML researchosphere is considerably difficult due to the widespread assumption that "every positive result is scientifically more valuable, or interesting, than any negative one". This is likely even more the case in DL research because of the ever-increasing competition. For instance, the percentage of accepted papers related to deep neural networks in the \textit{Conference on Computer Vision and Pattern Recognition (CVPR)}, one of the most prestigious in its field, has been 1\%, 14\% and 25\% for the years 2013, 2015 and 2017, respectively~\cite{deepCVPRstats}. Note that the amount of publication submissions to conferences and journals are increasing every year as well, e.g., the number of submissions to \textit{Annual Conference on Neural Information Processing Systems (NeurIPS)} doubled from 2016 to 2018~\cite{confstats}. Similar trends can be expected to be observed in research funding or scholarship applications. A research proposal is more likely to get a positive review if it builds further on "encouraging results" from previous work. There have been incentives to discuss the importance of negative results and share them in ML research~\cite{giraud2010importance} such as the \textit{First Workshop on Negative Results in Computer Vision} in 2017 and we hope more actions towards this direction will be realized in the future.

Current outcome reporting bias in ML/DL research is generated both from the authors' side as a reluctance to report negative results as well as the journals' side in selecting the results worth publishing and it is not trivial to separate the extent of the two. Even in presence of a positive result, authors may not report the negative ones, thinking such reporting will devaluate their work. As stated by Nissen et al., even if authors’ behavior is the main contributor to publication bias (there is evidence supporting this in other fields~\cite{olson2002publication,franco2014publication}), they may simply be responding to the editorial preferences for positive results~\cite{nissen2016publication}. The lack of traditional hypothesis construction before conducting the experiments and the lack of expectation to do so, supports the incentive of avoiding reporting of negative results in ML/DL field.

There are several consequences of such allergy against negative results in deep learning research. First, it eventually creates a bias against disruptive innovative ideas and favors incremental tweaks on well-established methods. Secondly, when negative results are not reported or published, it is essentially more difficult to construct causality and elaborate on the phenomena behind the positive results. As in other aspects of life, after all, we learn from negative results as well as positive ones. Furthermore, it increases the waste of resources and efforts due to unnecessary (re-)implementation of methods that have been shown to be inferior but never reported. Finally, the probability of a negative result being caused simply because of poor implementation exhibits the potential of that work being influential once implemented properly.

The trend of starting from a solution (often somebody else's) instead of from the problem itself and HARKing after minor modifications can be changed by changing our paradigm of publication process. Hereby, we propose a \textit{results-blind} review process for ML/DL research:
\begin{itemize}
    \item A paper is submitted with a clear hypothesis accompanied with the design of experiments. The hypothesis can be based on extensive analysis of previous studies, mathematical theory with unambiguous assumptions and/or domain knowledge of the specific field.
    \item The paper gets peer reviewed, preferably double-blind, and the reviewers suggest modifications and improvements on the experimental methods.
    \item Once accepted, the experiments are run.
    \item The paper gets published regardless of the results with a comprehensive discussion section. 
\end{itemize}
This approach would increase the likelihood of the study to be informative and influential regardless of the outcome, not only in the case of positive results. Essentially, the review process will give more attention to the experimental design and the hypothesis behind the proposed methods, decreasing the incentive for HARKing significantly. Naturally, this will also encourage researchers to navigate outside the "marginal improvements over the previous SotA" thinking. Similar ideas have been discussed especially in the field of psychology~\cite{locascio2017results,hyman2017can,woznyj2018results}. Note that we do not claim that the abovementioned proposal is applicable for every machine learning research publication process, mostly due to the scarcity of high quality reviewers. Nevertheless, we believe such discussions are beneficial and may eventually lead to improvements that will decrease the prevalence of HARKing in ML/DL research.




\section{"In The Wild" Illusion} \label{sec4}
Numerous studies in the field of deep learning utilize publicly available annotated datasets for computer vision, natural language processing, audio analysis and various other tasks. Several of these datasets even include the phrase "In the Wild" in their name - an expression to convey the message that the dataset holds no constraints and is representative of real-world circumstances. Even though it is not stated explicitly, the main assumption behind using these datasets is that the observations belonging to these datasets are drawn from the same statistical distribution of all possible observations naturally occurring in real-world.

In 2011, Torralba and Efros proposed to examine dataset bias in twelve popular image datasets by observing if it is possible to train a machine learning model to identify the dataset a given image is selected from~\cite{torralba2011unbiased}. Considering the random guess accuracy is only $\frac{1}{12}\approx8\%$, the authors found that humans were able to perform at $>75\%$, while a simple support vector machine classifier performed at $39\%$. The authors furthermore demonstrated the inability to perform cross-dataset generalization, thereby highlighting how models trained on typical datasets actually \textit{overfit} and thus fail to generalize to other datasets yet alone to real-world settings.  

A similar problem of overfitting stems from the hyper-competitive nature of machine learning, where there is little incentive of trying to publish methods that have inferior performance compared to SotA on test datasets (see Section \ref{sec3}). Therefore, we can reasonably expect that effectively most research uses the test set as a validation set, rather than following the standard practice of defining a separate validation set from the training data. Recht et al. show this by creating a new test set for CIFAR10, a widely used image dataset, where they found that there was a significant drop in accuracy (4-15\%) from the old test set to the new test set when tested with several DL architectures~\cite{recht2018cifar}. In a more recent work, a similar phenomenon is also shown for the well-known ImageNet dataset, suggesting that the accuracy drops are caused by the models' inability to generalize to slightly "harder" images than those found in the original test sets~\cite{recht2019imagenet}.

From the HARKing perspective, formulating hypotheses that are specifically designed to account for the observed results for a specific sample of observations go hand in hand with overfitting and failure of generalization. Furthermore, the selected datasets to run the proposed experiments on have to be in parallel with the hypothesis. For instance, the well-known \textit{Labeled Faces in the Wild} dataset~\cite{huang2008labeled} contains images of famous people only, but have been used extensively to test hypotheses of face recognition or person identification in unconstrained settings. And from the implementation perspective, by splitting a dataset into training, validation, and testing sets, we invariably risk giving the false impression that because our model may perform well on the test dataset, it will also generalize to images found in real world applications. In both cases mentioned above (using biased datasets and/or overfitting to specific test sets), it can be argued that hypotheses testing is conditional on the dataset in question, and therefore to convince a reader that HARKing has not occurred, an author should always take great care to demonstrate the generalizability of new methods. Obviously, overfitting is a problem encountered in ML in general and is not specific to neural networks. However, considering:
\begin{enumerate}[label=(\roman*)]
\item feed-forward neural networks are universal function approximators (by \textit{Universal Approximation Theorem}) as well as convolutional networks, i.e., a single hidden layer network containing a finite number of neurons can approximate continuous functions with arbitrary precision~\cite{yarotsky2018universal,zhou2018universality}
\item the complexity of the computed function by a neural network grows \textit{exponentially} with its depth, i.e., for every additional hidden layer, one needs exponentially more parameters to express the same function with a shallower network~\cite{eldan2016power,raghu2017expressive}
\end{enumerate}
deep neural architectures are very likely to suffer from overfitting due to their expressive power.

\section{Automated Machine Learning} \label{sec5}
The traditional data science approach relies on many sequential tasks; i.e. data preprocessing and cleaning, feature engineering and selection, model selection and parameter tuning, postprocessing, and finally critical analysis of results. Often in practice, the human decision making processes in these tasks are inefficient (see Section \ref{sec2}) or based on heuristics. Furthermore, the combined complexity of these tasks often present an insurmountable barrier for non-experts, and thus automated machine learning (AutoML) is a topic that has become increasingly popular in recent years, promising to automate (at least parts) of this pipeline in order to improve efficiency of machine learning and accelerating research. 

Recently, the most popular AutoML task has focused extensively on neural architecture search (NAS)~\cite{zoph2016neural,real2017large,suganuma2017genetic,pham2018efficient,liu2018progressive,real2018regularized,zoph2018learning}, i.e., automating the design of neural network architectures for the search of architectures that are superior to hand-crafted ones. Several other AutoML tasks include automated hyper-parameter optimization~\cite{klein2016fast}, activation function search~\cite{ramachandran2017searching}, optimizer search~\cite{bello2017neural}, data augmentation policy search~\cite{cubuk2018autoaugment} or even search for better hardware utilization in heterogeneously distributed (mixture of CPUs and GPUs) computing environments~\cite{mirhoseini2017device}. The methods behind such \textit{meta-learning} approaches are mainly based on \textit{Bayesian optimization}~\cite{klein2016fast}, \textit{evolutionary algorithms}~\cite{suganuma2017genetic,real2018regularized} or more recently on \textit{reinforcement learning}~\cite{zoph2016neural,ramachandran2017searching,mirhoseini2017device}. Some of these methods are available both to the academy as well as to the industry as open source software or in the form of software-as-a-service.

These advancements not only help us discover better DL models and solutions in terms of quantitative metrics than hand-engineered ones, but also carry the possibility to transform the everyday working practices of machine learning researchers and practitioners. With AutoML, data scientists are expected to offload a significant portion of their routine work and focus on tasks that require a higher level thinking and creativity. However, certain issues have been raised related to AutoML approaches lately. For instance, Scuito et al. demonstrate that the search policies of state-of-the-art NAS techniques are no better than random policies~\cite{sciuto2019evaluating}. Similarly, Li and Talwalkar show that random search with early-stopping is a competitive NAS baseline on two benchmark tasks - one from computer vision and one from natural language processing~\cite{li2019random}. In addition, they discuss the reproducibility issues of published NAS results by elaborating on the necessity of having a tremendous amount of computation resources, lack of available source material/code and questionable robustness of published results~\cite{li2019random}.

Interestingly, the pursuit of simplifying machine learning development resulted in a significant increase in algorithmic complexity of AutoML methods including complicated training routines and architecture transformations~\cite{li2019random}. This complexity makes it more difficult to pinpoint which components of the found solution is crucial for high performance. In addition, considering the lack of ablation studies (the analysis of systematic removal of components or features of a model in order to identify which of them are the most relevant) in many works, AutoML field creates a dangerous ground for HARKing.

\section{The \texttt{Insert\_Adjective\_Here AI} Wave} \label{sec6}

\subsection{Ethical AI} \label{sec6-1}
Ethical issues regarding current developments in machine learning are perhaps much more critical than they currently perceived to be; as we already encounter ethically questionable decisions given by algorithms, sometimes unbeknownst to us. Examples include replacing faces and voices in videos~\cite{DeepFake}, detecting people using WiFi signals~\cite{ZhaoWifiPose}, deciding whose life to risk in an eminent accident~\cite{Contissa2017} and generating fake news~\cite{radford2018language}. In various scenarios, ML impacts decisions on legal and ethical issues as well such as insurance, hiring, lending. Therefore, it is crucial to develop models that are fair and unbiased regardless of the biases in the data~\cite{kusner2017count,madras2019fairness}. This issue has been recently emphasized even by the European Commission in their ethics guidelines report for AI by underlining the importance of paying attention to situations involving more vulnerable groups such as children, persons with disabilities or minorities, or to situations with asymmetries of power or information (e.g. employee-employer or business-consumer)~\cite{EC_Ethics_2019}.

With established industries (e.g. example firearms), it is common for the researchers and developers to leave the responsibility of ethics to entities that follow them (e.g. arms sellers and legislators). However, most AI-based systems have been much faster to deploy than conventional technology. Therefore, it is highly desirable for researchers to discuss ethical implications of their work and create a dialogue about them at the earliest possible stage. While selecting research topics that raise ethical issues itself serves this purpose, the desire to present good results might deter the discussion.

Another important ethical issue revolves around covert AI systems. A human should always know if she/he is interacting with a human being or a machine, and it is the responsibility of us that this is reliably achieved. As AI practitioners, we should ensure that humans are made aware of - or able to request and validate the fact that – they interact with an AI identity~\cite{EC_Ethics_2019}. Thus, hypothesis forming process should be clear and unambiguous, and should consider the possible use cases or implications as well. And in this pursuit, HARKing won't do.

\subsection{Human-centric AI} \label{sec6-2}
At the current stage, ML/DL algorithms are often designed as tools for defined domain experts, thus they need to address human needs and psychology in a realistic manner. To decrease the amount of HARKing, high-level domain experts should be incorporated to the study teams from the beginning as a collective intelligence of domain experts has considerable benefits and should be utilized whenever possible~\cite{barnett2019comparative}. This will lead to more successful forming of \textit{a priori} hypotheses and in the end should put pressure on scrutinizing results that do not support these hypotheses. Previously, worrying examples of failure in this have surfaced, where there has been only a limited input from the domain experts~\cite{ross2018ibm}. High-level expertise is especially relevant to create scientific hypotheses and should be differentiated from defining practical use-cases and training of AI, where a diverse spectrum of possible users should be affiliated to the project. 

HARKing is potentially a serious threat especially in AI-driven change in medical practice. This applies mostly to the effect of failing to report \textit{a priori} hypotheses that are unsupported by the current results~\cite{rubin2017does}. The algorithms that will be used in medicine typically need to be clinically validated in laborious and high-cost trials~\cite{topol2019high}. Suppressing hypotheses after the results are known can lead to wrongly planned clinical trials, as the background scientific literature (meta-analyses) is biased and this can lead to losing credibility in the eyes of physicians and decision makers, together with spending a huge amount of limited human and financial resources available to run these trials.

\subsection{Explainable, transparent and interpretable AI} \label{sec6-3}

Explainable artificial intelligence (XAI) is not only interesting as an academic curiosity; it is a necessity for the future. Developing explainable and transparent systems, as well as tools to measure transparency, is crucial for ethical AI development (see section \ref{sec6-1}). The main concept of XAI is centered around \textit{causal attribution} as it is in human nature to understand causality naturally. Having such causal explanations will provide substantial leap in reaching human-like perception of AI systems and anthropomorphism~\cite{Gunning2019a}. Explainable AI and model interpretability may be used in a synonymous manner. However, we think that \textit{explainability} may fall under the causality domain and \textit{interpretability} may belong to the mechanistic explanation of the algorithmic and model internals~\cite{lipton2018a}.

Recent deep learning algorithms provide high predictive performance but limited ways to provide {\it reasoning} on how an algorithm produces such level of high performance that exceeds human abilities~\cite{Ha2018a}. Even though there have been studies addressing this problem and proposing solutions~\cite{bach2015pixel,ribeiro2016a,selvaraju2017grad}, a common consensus on performing interpretation of ML and especially DL models has not been reached. In fact, even the definition of {\it interpretability} itself is not established, neither mathematically nor axiomatically in the literature~\cite{lipton2018a}. Furthermore, recent studies question the robustness and security of these interpretation methods (e.g. to adversarial attacks)~\cite{ghorbani2017interpretation}.

From HARKing perspective, one can relatively easily reverse engineer results to fit in a desired interpretation~\cite{ribeiro2016a,ghorbani2017interpretation,hall2018introduction}. To avoid such practices, interpretable algorithms should not be reversible, nor should they only provide interpretation depending upon algorithmic priors. In this regard, approaches aiming at more theoretical explanations of \textit{why} deep learning works, from learning theory to statistical physics~\cite{tishby2015a,suezen17a,martin2018a}, may be classified as \textit{true} XAI research. These approaches, rather than focusing only on interpretation of the mechanistic approaches after the results are known, aim at finding an \textit{ab-initio} technique, i.e., from the first-principles, to design a deep learning system without HARKing. Similarly, use of causal inference has recently been shown to be promising in understanding underlying mechanisms of deep learning systems~\cite{narendra2018exp} and if descriptive, causal modals can answer prediction, intervention and counterfactual questions~\cite{selvaraju2018grad}.

In terms of transparency, an interesting question is whether are we, as humans, required to know all the details about the AI capabilities of the equipment and sensors that surround us. This can be argued both ways; for example, we know virtually nothing about the abilities of human drivers that use the same highway as we do. But similar to what happened with established technology in automotive (like ABS and automatic transmission), we should be able to know the workings, accuracy stats, advantages and disadvantages of emerging AI technologies. This concept overlaps with abovementioned mechanistic interpretability issue and perception of human-like attributions.

\subsection{Reproducible AI} \label{sec6-4}
AI research is known and as a result appreciated for its significant contributions to open science (e.g. preprint archives), open source (e.g. code repositories, sharing of trained models etc.), open data and reproducible research paradigms. Yet, as a sub-field of computer science, it still shares a similar reproducibility crisis~\cite{donoho2009a,munafo2017manifesto,baker20161,gundersen2018reproducible,hutson2018artificial}. As Donoho et al. suggested, a computational research paper is merely an advertisement unless it is presented with an underlying code and data~\cite{donoho2009a}. We believe one of the reasons of this reproducibility crisis is HARKing. 

One essential contribution to this crisis in ML and especially in DL research is the lack of understanding of distinction between \textit{repeatability} and \textit{reproducibility}~\cite{plesser2018reproducibility}. We consider repeatability as the ability to recreate the results of a study/paper and reproducibility as the ability to reach the same conclusions despite the variations in the irrelevant components of the experiments~\cite{woods2018expanding}. Obviously, the role of hypothesizing driven by sound scientific methodology is essential in differentiating the two. As discussed in Section \ref{sec2}, competitive nature of the field and elevated pressure of achieving research and business outputs in a fast manner lead to hurried claims of reproducibility (often confused with repeatability) just like the hurried claims of SotA. Once this is coupled with the avoidance of reporting negative results or similar selective reporting (see Section \ref{sec3}), reproducibility crisis becomes inevitable.

It is important to acknowledge the initiatives for encouraging and increasing reproducibility in ML/DL research. For instance, in NIPS 2019, a \textit{reproducibility checklist} and a code submission policy is introduced, in which the code is expected to accompany the accepted papers. In \textit{AAAI Conference on Artificial Intelligence} in 2019, a workshop on reproducible AI has been held. Similarly, a workshop on reproducibility in ML was held in \textit{International Conference on Learning Representations (ICLR)} in 2019. Nevertheless, open questions remain such as "How can we measure reproducibility?", "What does it mean for a paper to have successful or unsuccessful replications?" or "What can the ML community learn from other fields?".

\subsection{Accountable AI} \label{sec6-5}
Accountability of algorithmic decision-making systems (e.g. credit scoring) has been under discussion as well as under implementation for decades especially from the regulatory and legal perspective. However, the rapid pace of AI developments and real-world applications of them, introduced circumstances in which high-stakes decisions with significant consequences for people and broader society are made by ML algorithms. One such potential impact is an \textit{accident} which can be, in this context, defined as an unintended and harmful behavior that emerges from poor design of real-world AI systems. Amodei et al. provides several concrete examples of such possible problems in AI safety including negative side effects (e.g. due to poorly designed objective functions), sensitivity to distributional shifts (the environment shifting away from the training environment) and reward hacking (the system gaming its objective function)~\cite{amodei2016concrete}.

Naturally, AI accountability is intertwined with explainability, reproducibility, fairness and human-centrism of design of these systems. Policies for demanding explanations of algorithmic decisions may help preventing negative consequences or may unintentionally hinder innovation while providing little meaningful protection, depending on their implementation and execution. For instance, European Union General Data Protection Regulation (GDPR)~\cite{voigt2017eu} introduced a potential accountability mechanism by \textit{right to explanation} since May 2018, but the concrete consequences are still yet to be observed. Regarding the role of reproducibility in accountability of AI systems, the fatal accident recently caused by an autonomous car (belonging to Uber) is a suitable example. The preliminary report released by the United States National Transport Safety Board stated that the self-driving system software misclassified the pedestrian and the system was not designed to alert the human operator under such emergency conditions~\cite{UberNTSBReport}. For the fair design of AI systems from the accountability perspective, the Gender Shades study~\cite{buolamwini2018gender} serves as an interesting example. In the study, biases present in commercial automated facial analysis algorithms are presented~\cite{buolamwini2018gender} and consequently, a recent study elaborated on the concept of \textit{actionable auditing} by investigating the impact of publicly naming biased performance results of commercial AI products~\cite{raji2019actionable}. Certain opportunities for hybrid models in which humans and machines interact (for explaining failures~\cite{nushi2018towards} or intervening operations~\cite{dann2018policy}) towards better AI accountability are also proposed in recent studies.

From the industry perspective, considering large companies and corporations entering an "AI race" in order to be the first to successfully employ AI in their domains, it is not surprising for accountability to take lower priority over invention and market leadership. But from the scientific methodology perspective, taking accountability of ML/DL models into account in the early stages of the research process, such as hypothesis forming, is imperative.

\subsection{Privacy-aware AI} \label{sec6-6}
Current implementations of ML algorithms require access to data, which essentially opens up potential security and privacy risks. Therefore, privacy-aware or privacy-preserving AI notion and several studies along this paradigm has been conducted, leading to influential concepts including \textit{federated learning} and \textit{differential privacy}~\cite{mcmahan2016communication,abadi2016deep}. With the use of \textit{homomorphic encryption}, deep learning model inference on encrypted data was shown to be possible with a little trade-off from accuracy as well~\cite{cheon2017homomorphic,boemer2018ngraph}. In addition, Shokri et al. introduced and elaborated on the concept called \textit{membership inference attack}, i.e., given a black-box machine learning model and a data record, determining whether this record was used as part of the model’s training dataset or not~\cite{shokri2017membership}. All these advancements are crucial to declare that several metrics are needed to assess and compare ML models and privacy preserving capability is one of them. For a good scientific conduct, our hypotheses on both the methods and impacts of our research should consider these concepts.

\section{Conclusion}

Hypothesizing after the results are known has been observed in several fields of research throughout the history and recently deep learning research exhibits several instances of it as well. In this work, we tried to give examples of HARKing in machine learning and especially in deep learning research. We elaborated on the reasons and consequences of this troubling trend by discussing overemphasis on single-metric model comparisons and benchmarks (Section \ref{sec2}), tendency to refrain from reporting negative results (Section \ref{sec3}), failure of generalization (Section \ref{sec4}) and automatic machine learning (Section \ref{sec5}). Finally, HARKing and importance of formulating an a priori hypothesis is reviewed from the perspective of ethical, human-centric, explainable, reproducible, accountable and privacy-preserving AI notions (Section \ref{sec6}).

We would like to emphasize the importance of discussions for achieving concrete reforms in the mentioned issues. Cultural change and legitimate interventions (such as the proposal in Section \ref{sec3}) in deep learning research should be encouraged by addressing these issues as much as we can in a constructive manner. As the aimed progress is a collaborative effort, researchers, practitioners, reviewers, editors, policy-makers, decision-makers, funding agencies, corporations and governmental entities need to act collectively. We believe that prevention of HARKing will help in engineering ethical, accountable, transparent, unbiased and scientifically superior deep learning solutions for the common good of the society we will be living in eventually. We also hope and believe that this work will stir discussions and debates, and will contribute towards that goal.


\bibliographystyle{unsrt}
\bibliography{references}

\begin{thebibliography}{10}

\bibitem{kerr1998harking}
Norbert~L Kerr.
\newblock Harking: Hypothesizing after the results are known.
\newblock {\em Personality and Social Psychology Review}, 2(3):196--217, 1998.

\bibitem{hempel1966}
Carl~G. Hempel.
\newblock {\em Philosophy of Natural Science}.
\newblock O'Reilly Media, Incorporated, 1966.

\bibitem{murphy2017harking}
Kevin~R Murphy and Herman Aguinis.
\newblock Harking: How badly can cherry-picking and question trolling produce
  bias in published results?
\newblock {\em Journal of Business and Psychology}, pages 1--17, 2017.

\bibitem{hitchcock2004prediction}
Christopher Hitchcock and Elliott Sober.
\newblock Prediction versus accommodation and the risk of overfitting.
\newblock {\em The British journal for the philosophy of science}, 55(1):1--34,
  2004.

\bibitem{rubin2017does}
Mark Rubin.
\newblock When does harking hurt? identifying when different types of
  undisclosed post hoc hypothesizing harm scientific progress.
\newblock {\em Review of General Psychology}, 21(4):308--320, 2017.

\bibitem{lecun2015deep}
Yann LeCun, Yoshua Bengio, and Geoffrey Hinton.
\newblock Deep learning.
\newblock {\em Nature}, 521(7553):436, 2015.

\bibitem{lipton2018troubling}
Zachary~C Lipton and Jacob Steinhardt.
\newblock Troubling trends in machine learning scholarship.
\newblock {\em arXiv preprint arXiv:1807.03341}, 2018.

\bibitem{EC_Communication_2019}
European Commission.
\newblock Building trust in human-centric artificial intelligence.
\newblock April 2019.

\bibitem{nuzzo2015fooling}
Regina Nuzzo.
\newblock Fooling ourselves.
\newblock {\em Nature}, 526(7572):182, 2015.

\bibitem{bailey2014a}
David~H Bailey, Jonathan Borwein, Marcos Lopez~de Prado, and Qiji~Jim Zhu.
\newblock Pseudo-mathematics and financial charlatanism: The effects of
  backtest overfitting on out-of-sample performance.
\newblock {\em Notices of the American Mathematical Society}, 61(5):458--471,
  2014.

\bibitem{keskar2016large}
Nitish~Shirish Keskar, Dheevatsa Mudigere, Jorge Nocedal, Mikhail Smelyanskiy,
  and Ping Tak~Peter Tang.
\newblock On large-batch training for deep learning: Generalization gap and
  sharp minima.
\newblock {\em arXiv preprint arXiv:1609.04836}, 2016.

\bibitem{hastie2009elements}
Trevor Hastie, Robert Tibshirani, and Jerome Friedman.
\newblock The elements of statistical learning: data mining, inference, and
  prediction, springer series in statistics, 2009.

\bibitem{cawley2010over}
Gavin~C Cawley and Nicola~LC Talbot.
\newblock On over-fitting in model selection and subsequent selection bias in
  performance evaluation.
\newblock {\em Journal of Machine Learning Research}, 11(Jul):2079--2107, 2010.

\bibitem{reimers2017reporting}
Nils Reimers and Iryna Gurevych.
\newblock Reporting score distributions makes a difference: Performance study
  of lstm-networks for sequence tagging.
\newblock {\em arXiv preprint arXiv:1707.09861}, 2017.

\bibitem{selvin1966data}
Hanan~C Selvin and Alan Stuart.
\newblock Data-dredging procedures in survey analysis.
\newblock {\em The American Statistician}, 20(3):20--23, 1966.

\bibitem{head2015extent}
Megan~L Head, Luke Holman, Rob Lanfear, Andrew~T Kahn, and Michael~D Jennions.
\newblock The extent and consequences of p-hacking in science.
\newblock {\em PLoS biology}, 13(3):e1002106, 2015.

\bibitem{olsson2019prevalence}
Anton Olsson-Collentine, Marcel~ALM van Assen, and Chris~HJ Hartgerink.
\newblock The prevalence of marginally significant results in psychology over
  time.
\newblock {\em Psychological science}, page 0956797619830326, 2019.

\bibitem{sterling1959publication}
Theodore~D Sterling.
\newblock Publication decisions and their possible effects on inferences drawn
  from tests of significance—or vice versa.
\newblock {\em Journal of the American statistical association},
  54(285):30--34, 1959.

\bibitem{song2000publication}
Fujian Song, A~Eastwood, Simon Gilbody, Lelia Duley, and A~Sutton.
\newblock Publication and related biases: a review.
\newblock {\em Health technology assessment}, 4(10), 2000.

\bibitem{easterbrook1991publication}
Phillipa~J Easterbrook, Ramana Gopalan, JA~Berlin, and David~R Matthews.
\newblock Publication bias in clinical research.
\newblock {\em The Lancet}, 337(8746):867--872, 1991.

\bibitem{moscati1994positive}
Ronald Moscati, Dietrich Jehle, David Ellis, Albert Fiorello, and Michael
  Landi.
\newblock Positive-outcome bias: comparison of emergency medicine and general
  medicine literatures.
\newblock {\em Academic emergency medicine}, 1(3):267--271, 1994.

\bibitem{sterling1995publication}
Theodore~D Sterling, Wilf~L Rosenbaum, and James~J Weinkam.
\newblock Publication decisions revisited: The effect of the outcome of
  statistical tests on the decision to publish and vice versa.
\newblock {\em The American Statistician}, 49(1):108--112, 1995.

\bibitem{matosin2014negativity}
Natalie Matosin, Elisabeth Frank, Martin Engel, Jeremy~S Lum, and Kelly~A
  Newell.
\newblock Negativity towards negative results: a discussion of the disconnect
  between scientific worth and scientific culture, 2014.

\bibitem{deepCVPRstats}
Cvpr statistics.
\newblock \url{http://jponttuset.cat/are-gans-the-new-deep/}, 2018.
\newblock Accessed: 2019-04-07.

\bibitem{confstats}
Acceptance rates and submission numbers for main machine learning conferences.
\newblock \url{https://github.com/lixin4ever/Conference-Acceptance-Rate}, 2019.
\newblock Accessed: 2019-04-07.

\bibitem{giraud2010importance}
Christophe~G Giraud-Carrier and Margaret~H Dunham.
\newblock On the importance of sharing negative results.
\newblock {\em SIGKDD explorations}, 12(2):3--4, 2010.

\bibitem{olson2002publication}
Carin~M Olson, Drummond Rennie, Deborah Cook, Kay Dickersin, Annette Flanagin,
  Joseph~W Hogan, Qi~Zhu, Jennifer Reiling, and Brian Pace.
\newblock Publication bias in editorial decision making.
\newblock {\em Jama}, 287(21):2825--2828, 2002.

\bibitem{franco2014publication}
Annie Franco, Neil Malhotra, and Gabor Simonovits.
\newblock Publication bias in the social sciences: Unlocking the file drawer.
\newblock {\em Science}, 345(6203):1502--1505, 2014.

\bibitem{nissen2016publication}
Silas~Boye Nissen, Tali Magidson, Kevin Gross, and Carl~T Bergstrom.
\newblock Publication bias and the canonization of false facts.
\newblock {\em Elife}, 5:e21451, 2016.

\bibitem{locascio2017results}
Joseph~J Locascio.
\newblock Results blind science publishing.
\newblock {\em Basic and applied social psychology}, 39(5):239--246, 2017.

\bibitem{hyman2017can}
Michael~R Hyman.
\newblock Can “results blind manuscript evaluation” assuage “publication
  bias”?
\newblock {\em Basic and applied social psychology}, 39(5):247--251, 2017.

\bibitem{woznyj2018results}
Haley~M Woznyj, Kelcie Grenier, Roxanne Ross, George~C Banks, and Steven~G
  Rogelberg.
\newblock Results-blind review: a masked crusader for science.
\newblock {\em European Journal of Work and Organizational Psychology},
  27(5):561--576, 2018.

\bibitem{torralba2011unbiased}
A~Torralba and AA~Efros.
\newblock Unbiased look at dataset bias.
\newblock In {\em Proceedings of the 2011 IEEE Conference on Computer Vision
  and Pattern Recognition}, pages 1521--1528. IEEE Computer Society, 2011.

\bibitem{recht2018cifar}
Benjamin Recht, Rebecca Roelofs, Ludwig Schmidt, and Vaishaal Shankar.
\newblock Do cifar-10 classifiers generalize to cifar-10?
\newblock {\em arXiv preprint arXiv:1806.00451}, 2018.

\bibitem{recht2019imagenet}
Benjamin Recht, Rebecca Roelofs, Ludwig Schmidt, and Vaishaal Shankar.
\newblock Do imagenet classifiers generalize to imagenet?
\newblock {\em arXiv preprint arXiv:1902.10811}, 2019.

\bibitem{huang2008labeled}
Gary~B Huang, Marwan Mattar, Tamara Berg, and Eric Learned-Miller.
\newblock Labeled faces in the wild: A database forstudying face recognition in
  unconstrained environments.
\newblock In {\em Workshop on faces in'Real-Life'Images: detection, alignment,
  and recognition}, 2008.

\bibitem{yarotsky2018universal}
Dmitry Yarotsky.
\newblock Universal approximations of invariant maps by neural networks.
\newblock {\em arXiv preprint arXiv:1804.10306}, 2018.

\bibitem{zhou2018universality}
Ding-Xuan Zhou.
\newblock Universality of deep convolutional neural networks.
\newblock {\em arXiv preprint arXiv:1805.10769}, 2018.

\bibitem{eldan2016power}
Ronen Eldan and Ohad Shamir.
\newblock The power of depth for feedforward neural networks.
\newblock In {\em Conference on learning theory}, pages 907--940, 2016.

\bibitem{raghu2017expressive}
Maithra Raghu, Ben Poole, Jon Kleinberg, Surya Ganguli, and Jascha~Sohl
  Dickstein.
\newblock On the expressive power of deep neural networks.
\newblock In {\em Proceedings of the 34th International Conference on Machine
  Learning-Volume 70}, pages 2847--2854. JMLR. org, 2017.

\bibitem{zoph2016neural}
Barret Zoph and Quoc~V Le.
\newblock Neural architecture search with reinforcement learning.
\newblock {\em arXiv preprint arXiv:1611.01578}, 2016.

\bibitem{real2017large}
Esteban Real, Sherry Moore, Andrew Selle, Saurabh Saxena, Yutaka~Leon Suematsu,
  Jie Tan, Quoc Le, and Alex Kurakin.
\newblock Large-scale evolution of image classifiers.
\newblock {\em arXiv preprint arXiv:1703.01041}, 2017.

\bibitem{suganuma2017genetic}
Masanori Suganuma, Shinichi Shirakawa, and Tomoharu Nagao.
\newblock A genetic programming approach to designing convolutional neural
  network architectures.
\newblock In {\em Proceedings of the Genetic and Evolutionary Computation
  Conference}, pages 497--504. ACM, 2017.

\bibitem{pham2018efficient}
Hieu Pham, Melody~Y Guan, Barret Zoph, Quoc~V Le, and Jeff Dean.
\newblock Efficient neural architecture search via parameter sharing.
\newblock {\em arXiv preprint arXiv:1802.03268}, 2018.

\bibitem{liu2018progressive}
Chenxi Liu, Barret Zoph, Maxim Neumann, Jonathon Shlens, Wei Hua, Li-Jia Li,
  Li~Fei-Fei, Alan Yuille, Jonathan Huang, and Kevin Murphy.
\newblock Progressive neural architecture search.
\newblock In {\em Proceedings of the European Conference on Computer Vision
  (ECCV)}, pages 19--34, 2018.

\bibitem{real2018regularized}
Esteban Real, Alok Aggarwal, Yanping Huang, and Quoc~V Le.
\newblock Regularized evolution for image classifier architecture search.
\newblock {\em arXiv preprint arXiv:1802.01548}, 2018.

\bibitem{zoph2018learning}
Barret Zoph, Vijay Vasudevan, Jonathon Shlens, and Quoc~V Le.
\newblock Learning transferable architectures for scalable image recognition.
\newblock In {\em Proceedings of the IEEE conference on computer vision and
  pattern recognition}, pages 8697--8710, 2018.

\bibitem{klein2016fast}
Aaron Klein, Stefan Falkner, Simon Bartels, Philipp Hennig, and Frank Hutter.
\newblock Fast bayesian optimization of machine learning hyperparameters on
  large datasets.
\newblock {\em arXiv preprint arXiv:1605.07079}, 2016.

\bibitem{ramachandran2017searching}
Prajit Ramachandran, Barret Zoph, and Quoc~V Le.
\newblock Searching for activation functions.
\newblock {\em arXiv preprint arXiv:1710.05941}, 2017.

\bibitem{bello2017neural}
Irwan Bello, Barret Zoph, Vijay Vasudevan, and Quoc~V Le.
\newblock Neural optimizer search with reinforcement learning.
\newblock {\em arXiv preprint arXiv:1709.07417}, 2017.

\bibitem{cubuk2018autoaugment}
Ekin~D Cubuk, Barret Zoph, Dandelion Mane, Vijay Vasudevan, and Quoc~V Le.
\newblock Autoaugment: Learning augmentation policies from data.
\newblock {\em arXiv preprint arXiv:1805.09501}, 2018.

\bibitem{mirhoseini2017device}
Azalia Mirhoseini, Hieu Pham, Quoc~V Le, Benoit Steiner, Rasmus Larsen, Yuefeng
  Zhou, Naveen Kumar, Mohammad Norouzi, Samy Bengio, and Jeff Dean.
\newblock Device placement optimization with reinforcement learning.
\newblock {\em arXiv preprint arXiv:1706.04972}, 2017.

\bibitem{sciuto2019evaluating}
Christian Sciuto, Kaicheng Yu, Martin Jaggi, Claudiu Musat, and Mathieu
  Salzmann.
\newblock Evaluating the search phase of neural architecture search.
\newblock {\em arXiv preprint arXiv:1902.08142}, 2019.

\bibitem{li2019random}
Liam Li and Ameet Talwalkar.
\newblock Random search and reproducibility for neural architecture search.
\newblock {\em arXiv preprint arXiv:1902.07638}, 2019.

\bibitem{DeepFake}
Deepfake - an introduction.
\newblock \url{https://www.scip.ch/en/?labs.20181004}, 2018.
\newblock Accessed: 2019-03-04.

\bibitem{ZhaoWifiPose}
Mingmin Zhao, Tianhong Li, Mohammad Abu~Alsheikh, Yonglong Tian, Hang Zhao,
  Antonio Torralba, and Dina Katabi.
\newblock Through-wall human pose estimation using radio signals.
\newblock In {\em Proceedings of the IEEE Conference on Computer Vision and
  Pattern Recognition}, pages 7356--7365, 2018.

\bibitem{Contissa2017}
Giuseppe Contissa, Francesca Lagioia, and Giovanni Sartor.
\newblock The ethical knob: ethically-customisable automated vehicles and the
  law.
\newblock {\em Artificial Intelligence and Law}, 25(3):365--378, Sep 2017.

\bibitem{radford2018language}
Alec Radford, Jeffrey Wu, Rewon Child, David Luan, Dario Amodei, and Ilya
  Sutskever.
\newblock Language models are unsupervised multitask learners.
\newblock Technical report, Technical report, OpenAi, 2018.

\bibitem{kusner2017count}
Matt~J. Kusner, Joshua~R. Loftus, Chris Russell, and Ricardo Silva.
\newblock Counterfactual fairness.
\newblock In {\em Advances in Neural Information Processing Systems 30: Annual
  Conference on Neural Information Processing Systems 2017, 4-9 December 2017,
  Long Beach, CA, {USA}}, pages 4069--4079, 2017.

\bibitem{madras2019fairness}
David Madras, Elliot Creager, Toniann Pitassi, and Richard Zemel.
\newblock Fairness through causal awareness: Learning causal latent-variable
  models for biased data.
\newblock In {\em Proceedings of the Conference on Fairness, Accountability,
  and Transparency}, pages 349--358. ACM, 2019.

\bibitem{EC_Ethics_2019}
European Commission.
\newblock Ethics guidelines for trustworthy ai.
\newblock April 2019.

\bibitem{barnett2019comparative}
Michael~L Barnett, Dhruv Boddupalli, Shantanu Nundy, and David~W Bates.
\newblock Comparative accuracy of diagnosis by collective intelligence of
  multiple physicians vs individual physicians.
\newblock {\em JAMA network open}, 2(3):e190096--e190096, 2019.

\bibitem{ross2018ibm}
Casey Ross and Ike Swetlitz.
\newblock Ibm’s watson supercomputer recommended ‘unsafe and
  incorrect’cancer treatments, internal documents show.
\newblock {\em Stat News https://www. statnews.
  com/2018/07/25/ibm-watson-recommended-unsafe-incorrect-treatments}, 2018.

\bibitem{topol2019high}
Eric~J Topol.
\newblock High-performance medicine: the convergence of human and artificial
  intelligence.
\newblock {\em Nature Medicine}, 25(1):44, 2019.

\bibitem{Gunning2019a}
David Gunning.
\newblock Darpa's explainable artificial intelligence (xai) program.
\newblock In {\em Proceedings of the 24th International Conference on
  Intelligent User Interfaces}, IUI '19, pages ii--ii, New York, NY, USA, 2019.
  ACM.

\bibitem{lipton2018a}
Zachary~C. Lipton.
\newblock The mythos of model interpretability.
\newblock {\em Queue}, 16(3):30:31--30:57, June 2018.

\bibitem{Ha2018a}
Taehyun Ha, Sangwon Lee, and Sangyeon Kim.
\newblock Designing explainability of an artificial intelligence system.
\newblock In {\em Proceedings of the Technology, Mind, and Society},
  TechMindSociety '18, pages 14:1--14:1, New York, NY, USA, 2018. ACM.

\bibitem{bach2015pixel}
Sebastian Bach, Alexander Binder, Gr{\'e}goire Montavon, Frederick Klauschen,
  Klaus-Robert M{\"u}ller, and Wojciech Samek.
\newblock On pixel-wise explanations for non-linear classifier decisions by
  layer-wise relevance propagation.
\newblock {\em PloS one}, 10(7):e0130140, 2015.

\bibitem{ribeiro2016a}
Marco~Tulio Ribeiro, Sameer Singh, and Carlos Guestrin.
\newblock "why should i trust you?": Explaining the predictions of any
  classifier.
\newblock In {\em Proceedings of the 22Nd ACM SIGKDD International Conference
  on Knowledge Discovery and Data Mining}, KDD '16, pages 1135--1144, New York,
  NY, USA, 2016. ACM.

\bibitem{selvaraju2017grad}
Ramprasaath~R Selvaraju, Michael Cogswell, Abhishek Das, Ramakrishna Vedantam,
  Devi Parikh, and Dhruv Batra.
\newblock Grad-cam: Visual explanations from deep networks via gradient-based
  localization.
\newblock In {\em Proceedings of the IEEE International Conference on Computer
  Vision}, pages 618--626, 2017.

\bibitem{ghorbani2017interpretation}
Amirata Ghorbani, Abubakar Abid, and James Zou.
\newblock Interpretation of neural networks is fragile.
\newblock {\em arXiv preprint arXiv:1710.10547}, 2017.

\bibitem{hall2018introduction}
Patrick Hall and Navdeep Gill.
\newblock {\em Introduction to Machine Learning Interpretability}.
\newblock Oxford, England: Prentice-Hall, 2018.

\bibitem{tishby2015a}
N.~{Tishby} and N.~{Zaslavsky}.
\newblock Deep learning and the information bottleneck principle.
\newblock In {\em 2015 IEEE Information Theory Workshop (ITW)}, pages 1--5,
  April 2015.

\bibitem{suezen17a}
Mehmet S{\"{u}}zen, Cornelius Weber, and Joan~J. Cerd{\`{a}}.
\newblock Spectral ergodicity in deep learning architectures via surrogate
  random matrices.
\newblock {\em CoRR}, abs/1704.08303, 2017.

\bibitem{martin2018a}
Charles~H Martin and Michael~W Mahoney.
\newblock Implicit self-regularization in deep neural networks: Evidence from
  random matrix theory and implications for learning.
\newblock {\em arXiv preprint arXiv:1810.01075}, 2018.

\bibitem{narendra2018exp}
Tanmayee Narendra, Anush Sankaran, Deepak Vijaykeerthy, and Senthil Mani.
\newblock Explaining deep learning models using causal inference.
\newblock {\em CoRR}, abs/1811.04376, 2018.

\bibitem{selvaraju2018grad}
Ramprasaath~R. Selvaraju, Abhishek Das, Ramakrishna Vedantam, Michael Cogswell,
  Devi Parikh, and Dhruv Batra.
\newblock Grad-cam: Why did you say that? visual explanations from deep
  networks via gradient-based localization.
\newblock {\em CoRR}, abs/1610.02391, 2016.

\bibitem{donoho2009a}
David~L Donoho, Arian Maleki, Inam~Ur Rahman, Morteza Shahram, and Victoria
  Stodden.
\newblock Reproducible research in computational harmonic analysis.
\newblock {\em Computing in Science \& Engineering}, 11(1):8--18, 2009.

\bibitem{munafo2017manifesto}
Marcus~R Munaf{\`o}, Brian~A Nosek, Dorothy~VM Bishop, Katherine~S Button,
  Christopher~D Chambers, Nathalie~Percie Du~Sert, Uri Simonsohn, Eric-Jan
  Wagenmakers, Jennifer~J Ware, and John~PA Ioannidis.
\newblock A manifesto for reproducible science.
\newblock {\em Nature Human Behaviour}, 1(1):0021, 2017.

\bibitem{baker20161}
Monya Baker.
\newblock 1,500 scientists lift the lid on reproducibility.
\newblock {\em Nature News}, 533(7604):452, 2016.

\bibitem{gundersen2018reproducible}
Odd~Erik Gundersen, Yolanda Gil, and David~W Aha.
\newblock On reproducible ai: Towards reproducible research, open science, and
  digital scholarship in ai publications.
\newblock {\em AI Magazine}, 39(3), 2018.

\bibitem{hutson2018artificial}
Matthew Hutson.
\newblock Artificial intelligence faces reproducibility crisis, 2018.

\bibitem{plesser2018reproducibility}
Hans~E Plesser.
\newblock Reproducibility vs. replicability: a brief history of a confused
  terminology.
\newblock {\em Frontiers in neuroinformatics}, 11:76, 2018.

\bibitem{woods2018expanding}
Bronwyn Woods.
\newblock Expanding search in the space of empirical ml.
\newblock {\em arXiv preprint arXiv:1812.01495}, 2018.

\bibitem{amodei2016concrete}
Dario Amodei, Chris Olah, Jacob Steinhardt, Paul Christiano, John Schulman, and
  Dan Man{\'e}.
\newblock Concrete problems in ai safety.
\newblock {\em arXiv preprint arXiv:1606.06565}, 2016.

\bibitem{voigt2017eu}
Paul Voigt and Axel Von~dem Bussche.
\newblock The eu general data protection regulation (gdpr).
\newblock {\em A Practical Guide, 1st Ed., Cham: Springer International
  Publishing}, 2017.

\bibitem{UberNTSBReport}
Preliminary report highway hwy18mh010, by the united states transportation
  security board.
\newblock
  \url{https://www.ntsb.gov/investigations/AccidentReports/Reports/HWY18MH010-prelim.pdf},
  2018.
\newblock Accessed: 2019-03-04.

\bibitem{buolamwini2018gender}
Joy Buolamwini and Timnit Gebru.
\newblock Gender shades: Intersectional accuracy disparities in commercial
  gender classification.
\newblock In {\em Conference on Fairness, Accountability and Transparency},
  pages 77--91, 2018.

\bibitem{raji2019actionable}
Inioluwa~Deborah Raji and Joy Buolamwini.
\newblock Actionable auditing: Investigating the impact of publicly naming
  biased performance results of commercial ai products.
\newblock In {\em AAAI/ACM Conf. on AI Ethics and Society}, 2019.

\bibitem{nushi2018towards}
Besmira Nushi, Ece Kamar, and Eric Horvitz.
\newblock Towards accountable ai: Hybrid human-machine analyses for
  characterizing system failure.
\newblock In {\em Sixth AAAI Conference on Human Computation and
  Crowdsourcing}, 2018.

\bibitem{dann2018policy}
Christoph Dann, Lihong Li, Wei Wei, and Emma Brunskill.
\newblock Policy certificates: Towards accountable reinforcement learning.
\newblock {\em arXiv preprint arXiv:1811.03056}, 2018.

\bibitem{mcmahan2016communication}
H~Brendan McMahan, Eider Moore, Daniel Ramage, Seth Hampson, et~al.
\newblock Communication-efficient learning of deep networks from decentralized
  data.
\newblock {\em arXiv preprint arXiv:1602.05629}, 2016.

\bibitem{abadi2016deep}
Martin Abadi, Andy Chu, Ian Goodfellow, H~Brendan McMahan, Ilya Mironov, Kunal
  Talwar, and Li~Zhang.
\newblock Deep learning with differential privacy.
\newblock In {\em Proceedings of the 2016 ACM SIGSAC Conference on Computer and
  Communications Security}, pages 308--318. ACM, 2016.

\bibitem{cheon2017homomorphic}
Jung~Hee Cheon, Andrey Kim, Miran Kim, and Yongsoo Song.
\newblock Homomorphic encryption for arithmetic of approximate numbers.
\newblock In {\em International Conference on the Theory and Application of
  Cryptology and Information Security}, pages 409--437. Springer, 2017.

\bibitem{boemer2018ngraph}
Fabian Boemer, Yixing Lao, and Casimir Wierzynski.
\newblock ngraph-he: A graph compiler for deep learning on homomorphically
  encrypted data.
\newblock {\em arXiv preprint arXiv:1810.10121}, 2018.

\bibitem{shokri2017membership}
Reza Shokri, Marco Stronati, Congzheng Song, and Vitaly Shmatikov.
\newblock Membership inference attacks against machine learning models.
\newblock In {\em Security and Privacy (SP), 2017 IEEE Symposium on}, pages
  3--18. IEEE, 2017.

\end{thebibliography}

\end{document}